\documentclass{article}
\usepackage{aaai2026}
\usepackage[utf8]{inputenc}
\usepackage[T1]{fontenc}
\usepackage{lmodern}
\usepackage{microtype}
\usepackage{graphicx}
\usepackage{booktabs}
\usepackage{array}
\usepackage{tabularx}
\usepackage{makecell}
\usepackage{multirow}
\usepackage[sort&compress,sectionbib]{natbib}
\usepackage{xurl}
\usepackage{amsmath}
\usepackage{amssymb}
\usepackage{mathtools}
\usepackage{xspace}
\usepackage{placeins}
\usepackage{float}
\usepackage{subcaption}
\usepackage[dvipsnames]{xcolor}
\definecolor{LinkBlue}{HTML}{2B5B7C}
\usepackage[colorlinks=true,linkcolor=LinkBlue,citecolor=LinkBlue,urlcolor=LinkBlue,breaklinks=true]{hyperref}
\usepackage{cleveref}
\newcommand{\pio}{\ensuremath{\pi_{0.5}}\xspace}

\newcommand{\routervla}{\textsc{RouterVLA}\xspace}
\newcommand{\sr}{\mathrm{SR}}
\newcommand{\pp}{\,\mathrm{pp}}
\newcommand{\E}{\mathcal{E}}

\hypersetup{
  pdftitle={RouterVLA: Budgeted Commissioning and Expert Onboarding for Growing VLA Pools},
  pdfauthor={Xingyu Ren, Chugang Yi, Youran Sun},
  pdfsubject={Vision-language-action policy commissioning and expert onboarding},
  pdfkeywords={vision-language-action models, policy selection, commissioning, expert onboarding, robot learning}
}

\title{RouterVLA: Budgeted Commissioning and Expert Onboarding for Growing VLA Pools}

\author{
Xingyu Ren$^{1 *}$,
Chugang Yi$^{2}$\thanks{Equal contribution.},
Youran Sun$^{2}$\thanks{\href{mailto:syouran0508@gmail.com}{syouran0508@gmail.com}}
}

\affiliations{
$^1$The Chinese University of Hong Kong\\
$^2$University of Maryland, College Park
}

\begin{document}

\maketitle

\pagestyle{plain}
\thispagestyle{plain}

\begin{abstract}
Robotic teams often maintain several vision-language-action policies but still deploy one global winner.
We study two recurring decisions: which expert to deploy for a new condition and which candidate to add to the pool.
\routervla combines a split-clean prior and outcome-disjoint probes with onboarding that credits only failures the incumbent pool cannot handle.
Under an exactly cost-matched probe budget, it reaches 60.53\% held-out success, a $+1.64\pp$ gain over a semantic shortlist.
Both criteria independently converge on the same five experts, confirming that the candidates best positioned to cover the base pool's blind spots are also broadly capable.
\end{abstract}

\section{Introduction}

We consider a deployment workflow in which a robotics team runs a small number of qualification rollouts before choosing a policy.
A one-shot workflow consumes that evidence in the checkpoint decision and discards its condition-level structure.
We ask whether a deployment system can reuse that structure.

The question arises because modern vision-language-action (VLA) policy collections are heterogeneous rather than redundant.
Broad foundation models can be reliable on familiar scenes, while fine-tunes or compact specialists may dominate one skill family or perturbation.
LIBERO and LIBERO-Plus expose this conditional structure through task suites and controlled changes~\citep{liu2023libero,fei2025libero}.
One global winner cannot express it.

Model scaling continues to improve robot policies.
Recent generalist systems include RT-1, RT-2, Octo, OpenVLA, \pio, and related policies~\citep{brohan2022rt1,brohan2023rt2,padalkar2023openx,team2024octo,kim2024openvla,black2024pi0}.
Deployment also creates a pool-management problem.
An operational pool may contain foundation checkpoints, task-specific adapters, efficient fine-tunes, and policies from different groups.
Such a pool requires qualification, selection, and admission rules in addition to improvements in its individual policies.

\routervla studies this lifecycle through two linked decisions.
For a new task condition, the system may spend a small budget of probe rollouts before committing to one expert for a held-out execution.
For a new candidate policy, it must decide whether that policy warrants a place in the operational pool.
The first decision is \emph{budgeted commissioning}; the second is \emph{expert onboarding}.
They are complementary: a router cannot exploit an expert that onboarding did not admit, and a larger pool helps only when deployment recognizes when its specialists matter.

The first version of \routervla~\citep{ren2026routervla} showed that completed smoke-test profiles can support conditional expert selection in a fixed pool.
This update extends the lifecycle in three directions: budgeted commissioning (allocating a limited per-condition probe budget rather than requiring exhaustive profiles), expert onboarding (deciding which new candidates deserve admission into the pool), and routing--pool synergy analysis (showing that routing and pool expansion are complementary).
All experiments use a new unified split (seed 20260711) with split-clean semantic scores and priors.

\begin{figure*}[t]
  \centering
  \includegraphics[width=\textwidth]{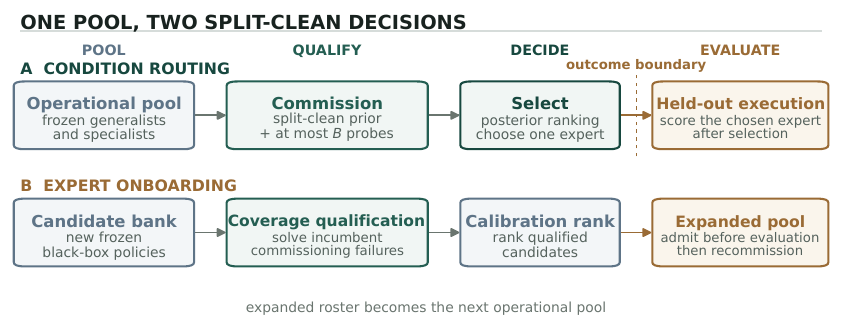}
  \caption{\textbf{RouterVLA treats policy qualification as a reusable lifecycle.}
  Split-clean historical evidence initializes a condition-specific ranking, outcome-disjoint probes refine it, and we score the chosen expert on a trial that never enters the router.
  A disjoint commissioning partition verifies that each candidate addition covers incumbent failures; calibration strength ranks the qualified candidates before evaluation scores the fixed expanded pool.}
  \label{fig:lifecycle}
\end{figure*}

The design in \cref{fig:lifecycle} uses a small set of fixed rules.
Routing starts from a prior that combines instruction similarity with leave-one-suite-out reliability estimated only from development conditions.
A short warm start tests plausible experts; the remaining budget goes to uncertain contenders close enough to the current leader to matter.
Onboarding first requires positive coverage of condition--trial failures left by the incumbent pool on the commissioning partition.
Calibration-mean success ranks the qualified candidates, and we fix five additions before opening the evaluation partition.
Marginal coverage and unfiltered calibration mean select the same set in this roster because its strongest candidates also cover incumbent failures.
The match makes no accuracy claim for coverage; coverage remains the rule that tests whether a candidate complements the incumbent pool.
A worst-suite floor provides a separate breadth diagnostic.

The primary routing comparison exactly matches probe cost at $B=6$.
At $B=6$, both \routervla and the semantic shortlist use six probes; they reach 60.53\% and 58.89\% success, respectively, for a paired improvement of $+1.64\pp$.
At the nominal $B=12$ cap, \routervla uses 12 probes while the fixed Top-3 semantic shortlist saturates at nine.
The $+2.88\pp$ gap is a budget-cap sensitivity rather than an exact cost match.
Separately, conditional deployment recovers $+13.53\pp$ more success than deploying a single global expert.
This contrast compares conditional deployment with one static expert; it is not a RouterVLA improvement over an equal-cost competitor.
Marginal coverage and calibration mean select the same five candidates in this roster.
The match shows only that these calibration-mean leaders also cover incumbent failures; it does not make the two criteria equivalent.
The selected 17-expert pool reaches 60.96\%, $12.16\pp$ above the mean of 20 random expansions and $0.01\pp$ above the full-roster point estimate.
The near-identical selected and full means do not constitute an equivalence test.

Our contributions are:
\begin{itemize}
  \item A split-clean, budget-aware commissioning rule for frozen heterogeneous VLA experts, evaluated with outcome-disjoint probes and held-out scoring.
  \item A cross-partition onboarding protocol that requires positive marginal coverage of incumbent failures, ranks qualified candidates by calibration mean, and never uses evaluation labels.
  \item A lifecycle study that separates the value of pool composition from the value of conditional routing as the roster expands.
\end{itemize}

\section{Related Work}

\paragraph{Generalist and specialist VLA policies.}
Large robot datasets and generalist policies have expanded the range of skills that one model can handle~\citep{brohan2022rt1,brohan2023rt2,padalkar2023openx,team2024octo}.
OpenVLA and OpenVLA-OFT make a large open VLA family practical to adapt~\citep{kim2024openvla,kim2025openvlaoft}.
Flow-matching and diffusion policies such as \pio and RDT broaden the range of VLA architectures~\citep{black2024pi0,liu2024rdt}.
Compact, modular, and suite-specialized systems including SmolVLA, LingBot-VLA, CogACT, MergeVLA, VLA-Adapter, X-VLA, and LaST-R1 create policy pools whose members are not interchangeable~\citep{shukor2025smolvla,wu2026pragmatic,li2024cogact,fu2025mergevla,wang2025vlaadapter,zheng2025xvla,chen2026lastr1}.
\routervla leaves these policies frozen and studies the deployment system built around them.

\paragraph{Conditional routing and algorithm selection.}
Choosing an algorithm conditionally is a classical problem~\citep{rice1976algorithm}; portfolio selectors such as SATzilla show how instance features can identify complementary black-box solvers rather than merely recover one global winner~\citep{xu2008satzilla}.
Recent systems route language models from preference data~\citep{ong2024routellm}, retrieve robotic policies from semantically similar executions~\citep{chen2026roborouter}, or route specialists inside modular VLA architectures~\citep{kuzmenko2025moira,fu2025mergevla}.
Our setting differs because the candidates are heterogeneous black-box policies and the router may actively purchase a small number of target-condition executions before deployment.

\paragraph{Budgeted evaluation.}
Fixed-budget best-arm identification separates a finite exploration phase from the final recommendation, while upper-confidence methods, Thompson sampling, successive halving, and Hyperband allocate finite evaluation budgets toward promising candidates~\citep{audibert2010best,auer2002ucb,russo2018thompson,jamieson2016successive,li2018hyperband}.
AutoML systems similarly treat evaluation as a scarce resource~\citep{thornton2013autoweka,feurer2015autosklearn}.
We do not propose a new generic bandit objective.
Instead, we contribute a robotics-specific prior, a trial-disjoint replay protocol, and a lifecycle that connects active selection with expert onboarding.

\section{RouterVLA}

\subsection{Budgeted Commissioning}

A condition $x$ specifies a task and a controlled perturbation, and $\E_x$ is the set of experts with valid outcomes for that condition.
The system may execute at most $B$ probe rollouts before selecting one expert $\hat e(x)$ for a scored rollout.
For every decision, we exclude the scored trial from all priors, probe histories, posteriors, and stopping decisions.

\paragraph{Split-clean prior.}
Task semantics identify plausible specialists, while historical reliability guards against a semantically attractive but brittle match.
Let $m_{x,e}\in[0,1]$ be a similarity-weighted success estimate from the ten nearest development conditions in all-MiniLM-L6-v2 embedding space~\citep{reimers2019sbert}.
Let $r_{s(x),e}\in[0,1]$ denote expert $e$'s development reliability; we estimate it while excluding the target suite $s(x)$.
We combine them as
\begin{equation}
q_{x,e}=0.10\,r_{s(x),e}+0.90\,m_{x,e}.
\label{eq:prior}
\end{equation}
We fix the mixing weight on the calibration partition, then refit both statistics on all development conditions before final evaluation.
Here ``calibrated'' describes mixture selection on held-out development data, not perfect probability calibration.

We use $q_{x,e}$ as the mean of a Beta prior.
After observing $c_e$ probe successes and $f_e$ failures, the posterior is
\begin{equation}
p_e\mid\mathcal D_x\sim\mathrm{Beta}\!\left(\tau_e q_{x,e}+c_e,\;\tau_e(1-q_{x,e})+f_e\right).
\label{eq:posterior}
\end{equation}
We clip $q_{x,e}$ to $[0.001,0.999]$ and set $\tau_e=2\max\{0.5,1-\operatorname{Var}_s(r_{s,e})\}$, so probe outcomes revise suite-unstable experts more strongly.
The deployed expert maximizes the posterior mean, with deterministic expert-ID tie breaking.

\paragraph{Budget-adaptive probes.}
At very small budgets, most probes should establish the semantic shortlist.
For $B\leq3$, the warm start uses $\max(1,B-1)$ prior-ranked probes; it uses two warm-start probes for $4\leq B\leq6$ and one for larger budgets.
The remaining probes prioritize experts that are both uncertain and close enough to the current leader to change the decision:
\begin{equation}
a_e=\frac{\sigma_e}{1+\max(\mu^*-\mu_e,0)},
\label{eq:acquisition}
\end{equation}
where $\mu_e$ and $\sigma_e$ are posterior moments and $\mu^*=\max_j\mu_j$.
No expert can consume more than the three non-scored trials available in the ledger.

\subsection{Expert Onboarding}

The 12-expert base pool follows the original \routervla configuration established in prior work~\citep{ren2026routervla}.
We treat it as an inherited operational pool rather than construct it for this evaluation.
The remaining 16 experts are candidate additions.
Let $\mathcal U(P)$ collect the commissioning condition--trial cells for which at least one expert in pool $P$ succeeds.
For each candidate $c$, we measure its marginal coverage against the fixed base pool $P_0$ and retain candidates with a positive gain:
\begin{equation}
\begin{aligned}
g_c&=\left|\mathcal U(P_0\cup\{c\})\setminus\mathcal U(P_0)\right|,\\
\mathcal C_+&=\{c:g_c>0\},\\
A_5&=\operatorname{TopK}_{c\in\mathcal C_+}\bar r_c,\\
P_{\mathrm{sel}}&=P_0\cup A_5,
\end{aligned}
\label{eq:onboarding}
\end{equation}
where $\bar r_c$ is candidate $c$'s mean success on the calibration partition; we predeclare $K=5$.
Expert ID breaks ties.
This coverage-qualified strength rule requires every addition to solve at least one incumbent blind spot before aggregate strength can rank it.
The two statistics answer different deployment questions: $\bar r_c$ measures a candidate's average strength, whereas $g_c$ measures complementarity relative to the incumbent pool.

We compare this operational criterion with two calibration rankings.
The first uses the same mean-success ranking but omits coverage qualification.
The unfiltered rule selects the same top-five set in this roster because every calibration-mean leader has $g_c>0$.
The match does not make the rules equivalent: unfiltered ranking can admit a high-mean candidate with $g_c=0$, whereas coverage qualification rejects it.
Let $\mathcal S$ contain the four benchmark suites.
The second diagnostic computes a breadth score $z_c=\min_{s\in\mathcal S}\widehat{\sr}_{s}(c)$.
This worst-suite statistic follows the same motivation as worst-group objectives~\citep{sagawa2020groupdro}, but here it is only a diagnostic and not a safety guarantee.
We use commissioning outcomes to fix $P_{\mathrm{sel}}$, then evaluate that frozen pool with the same six-probe router on the untouched evaluation partition.

\section{Experimental Design}

\paragraph{Benchmark and expert pool.}
Evaluating commissioning requires condition-level outcomes for many heterogeneous policies under shared perturbations.
The LIBERO-Plus perturbation ledger provides this structure across four suites and controlled camera, layout, and robot changes~\citep{fei2025libero}.
The study contains 398 task--perturbation conditions and 28 frozen expert IDs spanning generalists, fine-tunes, modular specialists, and compact policies.
The roster includes OpenVLA, OpenVLA-OFT, MergeVLA, RDT, CogACT, VLA-Adapter, X-VLA, SmolVLA, LingBot-VLA, and \pio-family policies~\citep{kim2024openvla,kim2025openvlaoft,fu2025mergevla,liu2024rdt,li2024cogact,wang2025vlaadapter,zheng2025xvla,shukor2025smolvla,wu2026pragmatic,black2024pi0}.
The source ledger contains 34{,}960 records.
We exclude 200 records from four checkpoints outside the declared roster, leaving 34{,}760 in-roster outcomes across 78.0\% of the nominal 398-by-28-by-4 study grid.
We deliberately preserve this observed availability pattern rather than impute missing outcomes.
Expert--task incompatibility, model crashes during source collection, and simulator initialization failures determine which cells exist.
That availability is part of commissioning because expert families differ in observation requirements, control interfaces, and executable condition coverage.

Before comparing methods, we fix one condition-specific validity mask for all of them.
The mask reports only whether the ledger contains a valid outcome; it neither reveals a valid cell's success label nor replaces an invalid cell.
Every relative comparison therefore uses the same candidate universe and information constraint.
We report absolute SR values conditional on this offline evaluation universe.
The complete roster appears in \cref{tab:expert-pool}.

\begin{figure*}[t]
  \centering
  \begin{minipage}[t]{0.54\textwidth}
    \centering
    \includegraphics[height=1.15in,keepaspectratio]{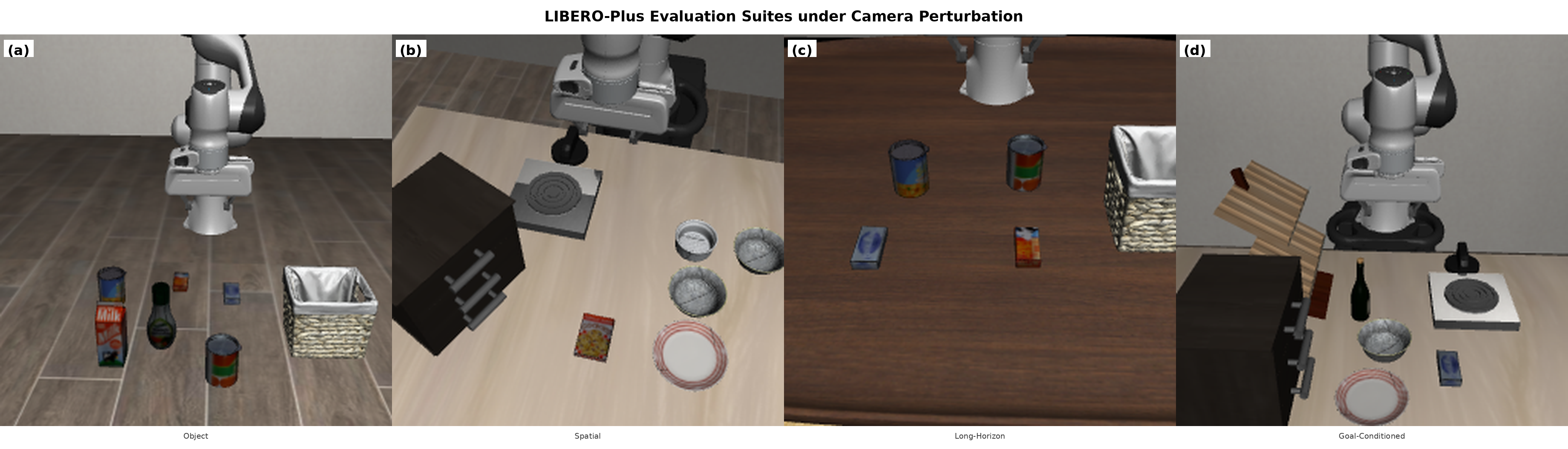}
  \end{minipage}
  \hfill
  \begin{minipage}[t]{0.44\textwidth}
    \centering
    \includegraphics[height=1.15in,keepaspectratio]{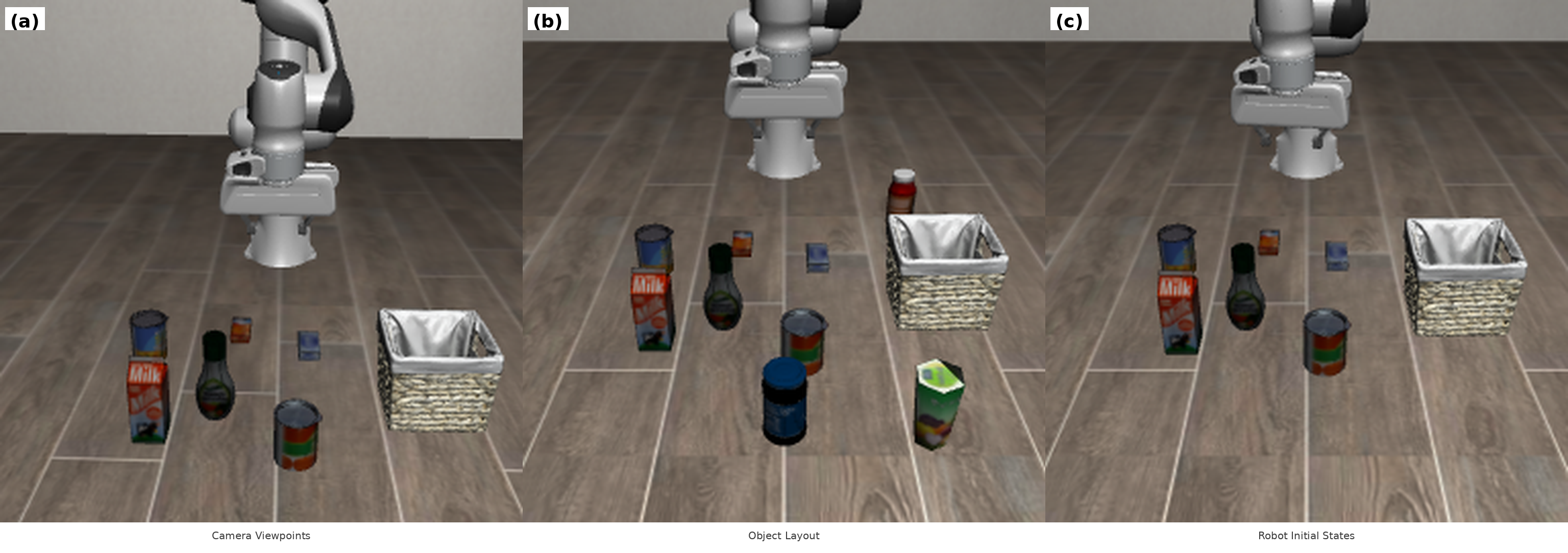}
  \end{minipage}
  \caption{\textbf{LIBERO-Plus evaluation context.}
  Representative simulator montages show the four task suites and perturbation axes.
  The images provide context; the router uses task text, development statistics, and qualification outcomes rather than image features.}
  \label{fig:libero}
\end{figure*}

\paragraph{Unified split.}
With seed 20260711, we stratify the 398 conditions into 80 calibration, 118 commissioning, and 200 evaluation conditions.
Stratification balances suite, development Global-Best success bins, and whether any available expert succeeds.
Stratification by success bin ensures representation across difficulty levels; we never use the evaluation partition for method decisions, so this binning does not expose outcomes to the router.
Calibration fixes the prior mixture and ranks onboarding candidates by mean success.
Commissioning determines which candidates pass positive marginal-coverage qualification, while evaluation supplies only final reports.
Our primary estimand is performance on held-out perturbation conditions from this benchmark catalog, not unseen task-template generalization.

\paragraph{Outcome-disjoint replay.}
Each valid condition--expert pair has four recorded simulator rollouts.
For every routing decision, we reserve one rollout for scoring and allow the other three to serve as probe outcomes; we repeat randomized acquisition ten times.
Replay does not synthesize policy behavior or fit a surrogate simulator.
It reuses recorded outcomes so every method faces the same candidates and counterfactual probe results while the router cannot see the scored execution.
This controlled protocol separates method differences from environmental noise and makes the probe--score boundary auditable.
Live deployment adds drift, interventions, and hardware variation; the present replay first establishes the comparison under identical outcomes.

\paragraph{Comparisons and uncertainty.}
The primary cost-matched baseline is a semantic Top-3 shortlist evaluated under the same probe budget as \routervla; at $B=6$, both methods use all six probes.
At the nominal $B=12$ cap, \routervla uses 12 probes.
The fixed three-expert shortlist exhausts its three available trials per expert and stops at nine.
We therefore treat that point as a budget-cap sensitivity rather than an exact cost match.
Secondary routing controls include a semantic-only hybrid using the same adaptive rule, reliability-only value-of-information (VOI), Top-M probing, Semantic Top-1 retrieval, Uniform Top-M, and random probing.
We list the zero-probe single global expert (Global Best) as a static deployment reference, not as a routing-method competitor.
Its contrast with conditional deployment compares one static expert with a routed pool rather than equal-cost methods.
Semantic Top-3 and Top-M probe a fixed prior-ranked shortlist; Uniform Top-M samples a shortlist uniformly.
Semantic Top-1 repeatedly tests its retrieved expert while trials remain.\footnote{This control isolates the semantic retrieval component of RoboRouter~\citep{chen2026roborouter}; it is not a reproduction of the full historical-memory system.}
At $B=6$, a matched gauntlet compares UCB, Thompson sampling, and successive halving under the same prior and a per-expert probe cap of 2.
For onboarding, we audit agreement between the coverage-qualified and unfiltered calibration-mean selections, compare the fixed selected pool with 20 random expansions, and report a worst-suite-floor diagnostic.
We report success rate (SR) and mean probes.
Primary 95\% confidence intervals use paired bootstrap resampling over 200 evaluation conditions; task-clustered and task-balanced sensitivities appear in the appendix.
\Cref{fig:libero,tab:protocol-summary} summarize the benchmark context and protocol.

\begin{table}[t]
\centering
\caption{Unified split and outcome-disjoint replay protocol used throughout the study.}
\label{tab:protocol-summary}
\footnotesize
\setlength{\tabcolsep}{2.6pt}
\renewcommand{\arraystretch}{1.04}
\begin{tabularx}{\columnwidth}{@{}X>{\raggedleft\arraybackslash}p{0.39\columnwidth}@{}}
\toprule
Quantity & Value \\
\midrule
Task--perturbation conditions & 398 \\
Frozen expert IDs & 28 \\
LIBERO suites & 4 \\
Recorded trials per valid condition--expert pair & 4 \\
Calibration / commissioning / evaluation & 80 / 118 / 200 \\
Primary probe budget & $B=6$ \\
Semantic neighbors / prior mixture & 10 / $\alpha=0.10$ \\
Scoring boundary & \makecell[r]{3 probe-eligible\\+ 1 held-out trial} \\
Replay repeats / bootstrap samples & 10 / 10{,}000 \\
Primary cluster & evaluation condition \\
\bottomrule
\end{tabularx}
\end{table}

\section{RouterVLA Improves Cost-Matched Routing}

\paragraph{Reliability mixing.}
The calibration partition selects the 0.10 reliability weight in \cref{eq:prior} from a fixed sweep.
Performance is nearly flat between weights 0 and 0.30, varying by less than $0.35\pp$, while larger reliability weights suppress the condition-specific semantic signal.
We therefore use the lightest nonzero correction that attains the highest calibration success and fix it before evaluation.

\begin{figure*}[t]
  \centering
  \includegraphics[width=\textwidth]{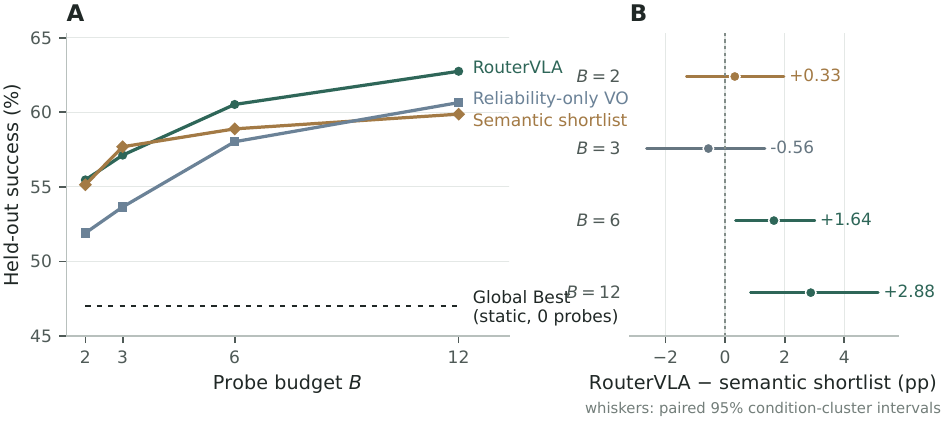}
  \caption{\textbf{Probe-budget sweep with an exact cost match at $B=6$.}
  \textbf{A:} Held-out success across nominal probe caps on the 200-condition evaluation partition.
  Global Best appears only as a zero-probe static deployment reference.
  \textbf{B:} Paired differences between RouterVLA and the semantic shortlist with 95\% condition-cluster intervals.
  At $B=12$, RouterVLA uses 12 probes while the shortlist saturates at nine, so this point is not an exact cost match.}
  \label{fig:routing}
\end{figure*}

\paragraph{Primary cost-matched comparison.}
\Cref{fig:routing,tab:main-routing} summarize the main routing result.
At the primary budget $B=6$, both \routervla and the semantic Top-3 shortlist use six probes.
They reach 60.53\% and 58.89\% held-out success, respectively.
The paired $+1.64\pp$ improvement has a 95\% condition-cluster interval of $[+0.35,+3.01]\pp$.
This shortlist is a strong baseline that already captures most of the condition-specific signal.
The $+1.64\pp$ is the additional value of reliability-aware adaptive commissioning after semantic selection and is the primary method comparison.
The reliability-adjusted hybrid (60.53\%) also improves over the semantic-only hybrid (58.98\%) by $+1.55\pp$, with interval $[+0.11,+3.30]\pp$.
This paired comparison isolates reliability mixing while holding adaptive probing fixed.
The remaining routing controls give 58.03\% for reliability-only VOI, 55.41\% for Top-M, and 53.38\% for Semantic Top-1 retrieval.

\paragraph{Conditional-deployment reference.}
Global Best deploys one expert across all conditions and uses no target-condition probes, so it is not a cost-matched router.
Conditional deployment recovers $+13.53\pp$ more success than this single global expert, which reaches 47.00\%.
This difference compares conditional deployment with one static expert rather than RouterVLA with an equal-cost method.

\Cref{tab:main-routing} also discloses a summary-only stopping diagnostic from a separate $B=6$ run.
Its same-prior confidence stop has the highest descriptive point estimate, 62.05\%, while using 5.69 probes on average.
The primary decision-level artifact and manifest do not contain matched records for this row, and the separate stopping run covers only $B=6$.
We report it descriptively but do not use it for a paired or headline claim.
The appendix records this row and the corresponding training-prior stopping diagnostic.

The budget sweep clarifies what reliability adjustment contributes.
At $B=2$, the reliability-adjusted and semantic methods differ by only $+0.33\pp$.
At $B=3$, the semantic shortlist is ahead by $0.56\pp$.
Once the budget reaches six probes, the policy can both establish plausible candidates and revisit uncertain alternatives, and the reliability-adjusted advantage becomes positive.
At the nominal $B=12$ cap, \routervla reaches 62.75\% with 12 probes and improves over the semantic shortlist, which reaches 59.88\% with nine probes, by $+2.88\pp$ with interval $[+0.85,+5.14]\pp$.
Because realized probe counts differ, this is a budget-cap sensitivity rather than a cost-matched method result.

\begin{table}[t]
\centering
\caption{Routing at $B=6$. The semantic shortlist is the primary cost-matched baseline; Global Best is a zero-probe static reference, and Same-prior stop is a separate diagnostic.}
\label{tab:main-routing}
\scriptsize
\setlength{\tabcolsep}{2.5pt}
\begin{tabular}{@{}lccc@{}}
\toprule
Method & SR & 95\% CI & Probes \\
\midrule
RouterVLA (primary) & \textbf{0.6052} & [0.5518, 0.6589] & 6.0 \\
Semantic shortlist (matched) & 0.5889 & [0.5324, 0.6434] & 6.0 \\
Semantic-only hybrid & 0.5898 & [0.5334, 0.6444] & 5.6 \\
Reliability-only VOI & 0.5803 & [0.5245, 0.6352] & 6.0 \\
Top-M & 0.5541 & [0.4973, 0.6106] & 6.0 \\
Semantic Top-1 retrieval & 0.5338 & [0.4775, 0.5900] & 3.0 \\
\midrule
Global Best (static) & 0.4700 & [0.4113, 0.5275] & 0.0 \\
\midrule
Same-prior stop$^\dagger$ & 0.6205 & [0.5648, 0.6751] & 5.7 \\
\bottomrule
\end{tabular}
\vspace{2pt}
\begin{minipage}{0.98\columnwidth}
\footnotesize
\raggedright
Paired 95\% condition-cluster intervals are $[+0.35,+3.01]\pp$ versus the semantic shortlist and $[+0.11,+3.30]\pp$ versus the semantic-only hybrid.
Conditional deployment recovers $+13.53\pp$ more success than Global Best.
Because Global Best uses zero probes, this compares conditional and static deployment rather than cost-matched methods.
$^\dagger$Aggregate-only result from a separate confidence-stopping run; the package lacks its matched decision records, so we make no paired claim for this row.
\end{minipage}
\end{table}

\paragraph{Uncertainty unit.}
The primary condition-cluster interval matches the deployment question in which a future perturbation condition receives equal weight.
Clustering the same paired decisions by task template gives $+1.64\pp$ with interval $[+0.34,+2.95]\pp$, and equal task weighting gives $+1.11\pp$ with interval $[+0.16,+2.29]\pp$.
A hierarchical task-balanced bootstrap is less precise at $B=6$, while all four estimands for the nominal-cap contrast have positive intervals at $B=12$.
These analyses support a condition-generalization claim without treating repeated trials as independent observations.

\begin{figure}[t]
  \centering
  \includegraphics[width=0.96\columnwidth]{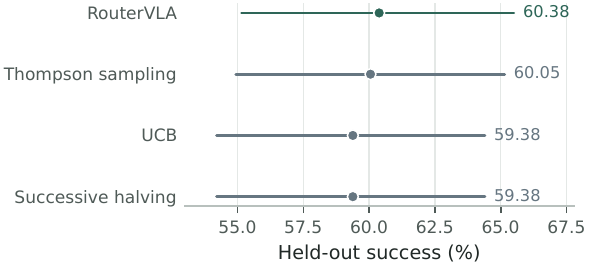}
  \caption{\textbf{Matched acquisition gauntlet at $B=6$.}
  All rules use the same prior and a per-expert probe cap of 2.
  RouterVLA has the highest mean, but the paired intervals do not resolve differences among the four active allocation rules.}
  \label{fig:gauntlet-main}
\end{figure}
\paragraph{Acquisition-rule sensitivity.}
No unusually strong generic bandit rule accounts for the gain.
In a matched $B=6$ gauntlet, \routervla obtains 60.38\%, Thompson sampling 60.05\%, and UCB and successive halving 59.38\% each.
The paired intervals among these four active allocation rules overlap zero, so we interpret the custom acquisition rule as competitive rather than universally superior.
The primary statistically resolved comparison is against the cost-matched semantic shortlist, which indicates that the split-clean prior and target-condition updates drive the result.
\Cref{fig:gauntlet-main,tab:acquisition-gauntlet} report the matched means and paired intervals.

\section{Onboarding New Experts}

Routing cannot use a specialist that is absent from the pool.
We next ask whether disjoint qualification evidence can identify candidates that add capability before we inspect the evaluation partition.

\begin{figure*}[t]
  \centering
  \includegraphics[width=\textwidth]{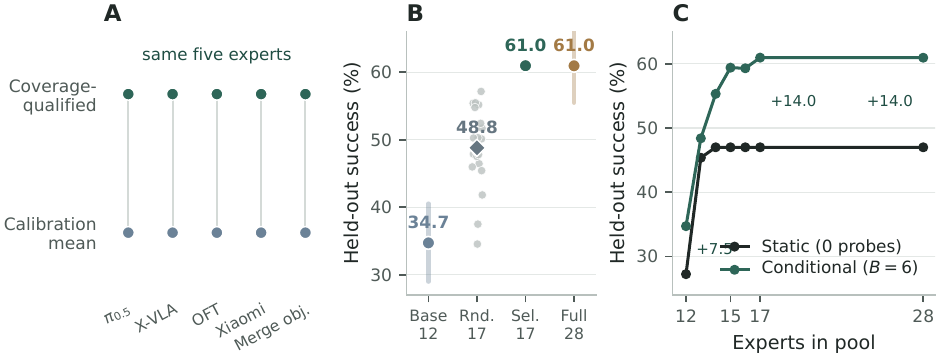}
  \caption{\textbf{Qualification evidence supports onboarding and deployment.}
  \textbf{A:} Marginal coverage and calibration mean select the same five experts in this roster, although only coverage tests complementarity to the base pool.
  \textbf{B:} The selected pool exceeds all 20 random expansions and nearly matches the full-roster point estimate; intervals appear only for the fixed base and full pools.
  \textbf{C:} Zero-probe static and six-probe conditional deployment along one fixed ordering of the selected set.}
  \label{fig:onboarding-lifecycle}
\end{figure*}

\subsection{Coverage Qualification and Strength}

The rule in \cref{eq:onboarding} first removes candidates that solve no commissioning cells missed by the base pool.
It then ranks the qualified candidates by calibration-mean success and selects \pio, X-VLA, OpenVLA-OFT, Xiaomi Robotics-0~\citep{cai2026xiaomi}, and MergeVLA object~\citep{fu2025mergevla}.

Marginal coverage and calibration-mean select the same five experts, but for different reasons — and that difference matters when the roster changes.
Calibration-mean never asks whether an expert complements the incumbent pool.
A candidate succeeding exclusively on conditions the base pool already solves would still rank highly.
Marginal coverage explicitly ignores those successes and only credits failures the base pool cannot handle.
In the current roster the two rankings coincide because the experts that add the most new coverage also happen to be strong.
But the decision rule produces a different selection as soon as a narrow specialist addresses a concentrated blind spot — a scenario our experiment is not powered to create but that a deployed system must handle.
We therefore present marginal coverage as a principled complementarity-aware criterion, not as an accuracy improvement over mean-success ranking in this particular roster.

MergeVLA object makes the complementarity concrete: it has 61.3\% success on the Object suite and near-zero success elsewhere, yet covers failures that broader experts leave behind.
The held-out pool result below evaluates the fixed set produced by this rule.
The minimum-suite-floor diagnostic ranks MergeVLA object eleventh and admits GR00T N1 instead.
We report that breadth diagnostic in the appendix, separate from the coverage-qualified pool used below.
Commissioning fixes the five-expert set, and only then do we score it on the untouched evaluation partition.

\subsection{Held-Out Pool Expansion}

Under six-probe routing, the 12-expert base pool reaches 34.73\% evaluation success.
Adding the five selected experts raises success to 60.96\%.
The 20 random five-expert expansions range from 34.55\% to 57.15\%, with mean 48.80\% and a between-draw standard deviation of $5.94\pp$.
The selected pool exceeds every random draw and is $12.16\pp$ above their mean.

The complete 28-expert roster reaches 60.95\%.
The selected 17-expert pool is $0.01\pp$ above that point estimate while maintaining 11 fewer experts.
This descriptive comparison is not an equivalence test.
\Cref{fig:onboarding-lifecycle,tab:onboarding-summary} report the pool comparisons.

\begin{table}[t]
\centering
\caption{Held-out routed success for five-expert pool expansions.}
\label{tab:onboarding-summary}
\scriptsize
\setlength{\tabcolsep}{2.5pt}
\begin{tabular}{@{}lccc@{}}
\toprule
Pool & Experts & SR & Fixed-pool 95\% CI \\
\midrule
Base & 12 & 0.3473 & [0.2900, 0.4053] \\
Random mean & 17 & 0.4880 & n/a \\
Selected & 17 & \textbf{0.6096} & n/a \\
Full roster & 28 & 0.6095 & [0.5550, 0.6625] \\
\bottomrule
\end{tabular}
\vspace{2pt}
\begin{minipage}{0.98\columnwidth}
\footnotesize
\raggedright
Fixed-pool CIs cluster conditions; random-pool SD is 0.0594 across 20 draws.
Marginal coverage and calibration mean select the same five experts in this roster.
Only coverage qualification tests whether a candidate solves an incumbent failure; we make no accuracy claim from the match.
The selected pool exceeds all 20 random draws and is $12.16\pp$ above their mean.
Its point estimate is $0.01\pp$ above the full roster; we do not infer equivalence from this unpaired comparison.
\end{minipage}
\end{table}

\section{Routing and Pool Growth Are Complementary}

Pool growth makes more capability available; routing decides where to deploy it.
Panel C of \cref{fig:onboarding-lifecycle} and \cref{tab:lifecycle-trajectory} trace both along one fixed ordering of the coverage-qualified set.

Along this ordering, routed success rises from 34.73\% to 48.40\% with \pio, 55.33\% with X-VLA, 59.40\% with OpenVLA-OFT, and 60.96\% after all five additions.
We fixed $K=5$ before evaluation, so the intermediate dip and plateau describe one path rather than a learned stopping rule or monotonic growth law.
The full roster reaches 60.95\%.

\begin{table}[t]
\centering
\caption{Static (zero-probe) and conditional ($B=6$) deployment along one fixed ordering of the selected set. The gain compares the two deployment rules.}
\label{tab:lifecycle-trajectory}
\footnotesize
\setlength{\tabcolsep}{3.0pt}
\begin{tabular}{@{}lrrrr@{}}
\toprule
Pool & Experts & Static & Routed & Gain (pp) \\
\midrule
Base & 12 & 0.2725 & 0.3473 & +7.47 \\
$+$ \pio & 13 & 0.4537 & 0.4840 & +3.02 \\
$+$ X-VLA & 14 & 0.4700 & 0.5533 & +8.33 \\
$+$ OpenVLA-OFT & 15 & 0.4700 & 0.5940 & +12.40 \\
$+$ Xiaomi Robotics-0 & 16 & 0.4700 & 0.5930 & +12.30 \\
$+$ MergeVLA object & 17 & 0.4700 & \textbf{0.6096} & \textbf{+13.96} \\
Full roster & 28 & 0.4700 & 0.6095 & +13.95 \\
\bottomrule
\end{tabular}
\end{table}

At 12, 17, and 28 experts, six-probe conditional deployment exceeds static deployment by $+7.48\pp$, $+13.96\pp$, and $+13.95\pp$, respectively.
These comparisons, together with the fixed trajectory, show complementary value from pool composition and routing without asserting a universal growth law.

\paragraph{Operational scale.}
Exhaustive qualification could consume 84 rollouts per condition; the primary operating point uses six and maintains 17 rather than 28 policies.
These are execution counts, not wall-clock latency.

\section{Discussion and Scope}

\paragraph{Operational pool size.}
The coverage-qualified pool reaches 60.96\% with 17 experts versus 60.95\% with all 28, maintaining 11 fewer policies.
Its $0.01\pp$ higher point estimate is descriptive, not evidence of equivalence.

\paragraph{Reading the routing effects.}
At $B=6$, both methods use six probes, making the paired $+1.64\pp$ over the 58.89\% semantic shortlist the primary cost-matched effect.
At the nominal $B=12$ cap, \routervla uses 12 probes versus nine for the shortlist, so the $+2.88\pp$ gap is budget-cap sensitivity rather than exact cost-matched evidence.
The separate $+13.53\pp$ gain compares conditional deployment with one zero-probe global expert.
The acquisition gauntlet does not resolve differences among the tested active rules, so we do not claim a universally superior bandit heuristic.

\paragraph{Why the onboarding criteria differ.}
Calibration mean measures aggregate strength; marginal coverage measures incumbent blind spots.
Both select the same five experts here because this roster's strongest candidates also add coverage.
We make no accuracy claim for coverage in this roster; its role is to exclude redundant candidates when the roster changes.
The commissioning partition fixes the admitted set without evaluation labels, and the worst-suite floor remains a separate breadth diagnostic.

\paragraph{Why controlled replay is the evaluation target.}
Across 28 frozen policies and controlled perturbations, replay gives every method identical candidates, probe outcomes, and scored executions, isolating the method comparison.
Physical deployment adds drift, wear, interventions, and latency; the present estimand deliberately targets controlled replay.

\paragraph{Shared availability mask.}
The ledger's 78.0\% grid coverage reflects differences in expert interfaces, executable conditions, and source-run stability.
One fixed validity mask gives every method the same candidate universe and reveals availability, not success labels.
It therefore defines the offline evaluation universe: absolute rates are conditional on it, and relative comparisons share the same information constraint.

\paragraph{Evidence boundaries.}
Routing priors use development conditions, commissioning fixes the onboarding set, and the final pool comparisons use untouched evaluation conditions.
The estimand covers held-out catalog perturbations, not unseen task templates; task-balanced analyses retain the direction with wider intervals.

\section{Conclusion: Cost-Matched Routing}

\routervla turns qualification into persistent deployment and admission decisions.
With six probes, its split-clean prior and outcome-disjoint updates reach 60.53\% success and improve over the exactly matched semantic shortlist by $+1.64\pp$.
At the nominal $B=12$ cap, \routervla uses 12 probes versus nine for the shortlist; the $+2.88\pp$ gap is budget-cap sensitivity, not equal-cost evidence.
The separate $+13.53\pp$ gain compares conditional deployment with one zero-probe global expert.
Marginal coverage and calibration mean choose the same five additions in the current roster.
Only marginal coverage asks whether a candidate complements the incumbent pool, so the match does not establish an accuracy gain for coverage.
Their 17-expert pool reaches 60.96\% versus 60.95\% for all 28, a descriptive point comparison rather than an equivalence result.
Under one fixed validity mask and controlled replay, policy admission and per-condition commissioning remain distinct, complementary deployment decisions.

\par\smallskip
\noindent\textbf{Acknowledgments.}
This work benefited from Agon~\citep{sun2026agon}, an autonomous research system built on Prompt Economy~\citep{sun2026perspectivegapbenchmarkmultiagentorchestration}.

\FloatBarrier
\clearpage

\phantomsection
\label{lp:bib-start}
\bibliography{references}

\clearpage
\appendix
\onecolumn
\section{Supplementary Material}
\label{sec:appendix}

This supplement provides commissioning pseudocode, the complete routing matrix, inference sensitivities, acquisition and stopping controls, candidate-level onboarding evidence, all random pool expansions, the full expert roster, and a self-contained restricted-pool specialist-coalition experiment.
The included CSV, JSON, and Parquet snapshots reproduce the values in the paper.

\subsection{Split and Trial-Disjoint Evaluation}

The unified split uses seed 20260711 and contains 80 calibration, 118 commissioning, and 200 evaluation conditions.
We stratify by LIBERO suite, development Global-Best success bin, and whether any candidate expert succeeds.
No condition appears in more than one partition.
Stratification by success bin ensures representation across difficulty levels; we never use the evaluation partition for method decisions, so this binning does not leak outcome information to the router.
The semantic retrieval score for an evaluation condition uses only the 198 development conditions, and suite reliability excludes the target suite.

The source ledger contains 34{,}760 valid in-roster records out of 44{,}576 nominal condition--expert--trial cells, a retention rate of 78.0\%.
We preserve this observed availability mask rather than impute missing outcomes.
Expert--task incompatibility, source-run crashes, and simulator initialization failures account for the missing cells.
Before any method comparison, we fix the mask and give every method the same valid candidate set for each condition.
The mask reports whether an outcome exists; it does not reveal the success of any valid outcome or impute a label for an invalid one.
Consequently, every method comparison uses identical replay inputs, while absolute success rates remain conditional on this offline evaluation universe.

\begin{figure}[H]
  \centering
  \includegraphics[width=0.86\textwidth]{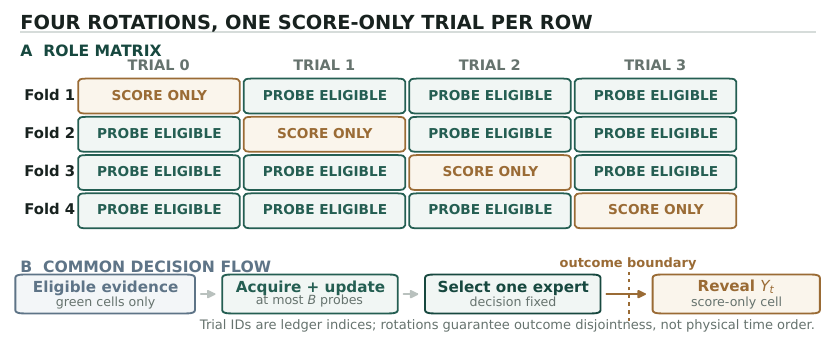}
  \caption{Outcome-disjoint evaluation.
  Each rotation reserves one recorded rollout for scoring and exposes the other three as probe-eligible outcomes.
  The scored trial never enters the semantic score, prior, acquisition sequence, posterior, or final selection.}
  \label{fig:protocol}
\end{figure}
\Cref{fig:protocol} shows the outcome boundary used in every replay decision.

\subsection{Commissioning Pseudocode}

\begin{figure}[H]
  \centering
  \setlength{\fboxsep}{7pt}
  \fbox{\begin{minipage}{0.94\textwidth}
  \textbf{Algorithm 1: Budgeted reliability-adjusted commissioning}\par
  \smallskip
  \textbf{Input:} condition $x$; probe budget $B$; nonempty eligible set $\E_x$ under the shared validity mask; development statistics $m_{x,e}$ and $r_{s,e}$; up to three non-scored outcomes per expert.\par
  \begin{enumerate}
    \item \textbf{Prior.} For every $e\in\E_x$, compute $q_{x,e}=0.10r_{s(x),e}+0.90m_{x,e}$, clip it to $[0.001,0.999]$, set $\tau_e=2\max\{0.5,1-\operatorname{Var}_s(r_{s,e})\}$, and initialize $(\alpha_e,\beta_e)=(\tau_e q_{x,e},\tau_e(1-q_{x,e}))$ and probe count $n_e=0$.
    \item \textbf{Warm start.} Set $h=\max(1,B-1)$ for $B\leq3$, $h=2$ for $4\leq B\leq6$, and $h=1$ for $B>6$.
    Probe once each of the first $\min(h,B)$ eligible experts in descending $q_{x,e}$ order and update its Beta counts.
    \item \textbf{Adaptive acquisition.} While the procedure has used fewer than $B$ probes, restrict attention to $A_x=\{e\in\E_x:n_e<3\}$.
    Stop if $A_x$ is empty.
    For $e\in A_x$, compute posterior mean $\mu_e$, standard deviation $\sigma_e$, leader mean $\mu^*=\max_j\mu_j$, and $a_e=\sigma_e/[1+\max(\mu^*-\mu_e,0)]$.
    Probe the expert maximizing $a_e$ and update $(\alpha_e,\beta_e,n_e)$.
    \item \textbf{Recommend.} Deploy $\hat e(x)=\arg\max_{e\in\E_x}\alpha_e/(\alpha_e+\beta_e)$ on the reserved scored trial.
  \end{enumerate}
  Expert ID breaks every tie in the prior order, acquisition score, and final recommendation.
  The reserved scored outcome is never eligible for an update.
  \end{minipage}}
  \caption{Complete commissioning procedure used in the replay experiments.}
  \label{alg:commissioning}
\end{figure}

\Cref{alg:commissioning} gives the complete fixed-budget procedure.
The four held-out-trial rotations and ten randomized probe-order repeats yield 8{,}000 routed decisions per method and budget on the 200 evaluation conditions.
Confidence intervals resample the evaluation condition as the primary independent unit.

\subsection{Complete Routing Results}

\begin{table}[H]
\centering
\caption{Routing results under shared nominal probe caps, followed by the zero-probe static deployment reference.}
\label{tab:full-routing-matrix}
\fontsize{7.2}{8.2}\selectfont
\setlength{\tabcolsep}{2.7pt}
\renewcommand{\arraystretch}{1.07}
\begin{tabular}{@{}lcccc@{}}
\toprule
Method & $B=2$ & $B=3$ & $B=6$ & $B=12$ \\
\midrule
\textbf{RouterVLA} & \textbf{\makecell{0.5546 [0.4989, 0.6086]\\2.0 probes}} & \textbf{\makecell{0.5713 [0.5159, 0.6260]\\3.0 probes}} & \textbf{\makecell{0.6053 [0.5518, 0.6589]\\6.0 probes}} & \textbf{\makecell{0.6275 [0.5706, 0.6825]\\12.0 probes}} \\
Semantic-only hybrid & \makecell{0.5454 [0.4901, 0.5995]\\1.9 probes} & \makecell{0.5710 [0.5165, 0.6259]\\2.9 probes} & \makecell{0.5898 [0.5334, 0.6444]\\5.6 probes} & \makecell{0.6090 [0.5536, 0.6636]\\10.6 probes} \\
Semantic shortlist & \makecell{0.5514 [0.4961, 0.6053]\\2.0 probes} & \makecell{0.5769 [0.5224, 0.6300]\\3.0 probes} & \makecell{0.5889 [0.5324, 0.6434]\\6.0 probes} & \makecell{0.5988 [0.5425, 0.6538]\\9.0 probes} \\
Reliability-only VOI & \makecell{0.5190 [0.4611, 0.5760]\\2.0 probes} & \makecell{0.5365 [0.4763, 0.5959]\\3.0 probes} & \makecell{0.5803 [0.5245, 0.6353]\\6.0 probes} & \makecell{0.6065 [0.5496, 0.6615]\\12.0 probes} \\
Top-M & \makecell{0.5190 [0.4623, 0.5766]\\2.0 probes} & \makecell{0.5353 [0.4794, 0.5904]\\3.0 probes} & \makecell{0.5541 [0.4973, 0.6106]\\6.0 probes} & \makecell{0.5675 [0.5088, 0.6250]\\9.0 probes} \\
Semantic Top-1 retrieval & \makecell{0.5338 [0.4775, 0.5888]\\2.0 probes} & \makecell{0.5338 [0.4787, 0.5900]\\3.0 probes} & \makecell{0.5338 [0.4775, 0.5900]\\3.0 probes} & \makecell{0.5338 [0.4775, 0.5900]\\3.0 probes} \\
Uniform Top-M & \makecell{0.2270 [0.1763, 0.2820]\\2.0 probes} & \makecell{0.2345 [0.1824, 0.2889]\\3.0 probes} & \makecell{0.2600 [0.2052, 0.3178]\\6.0 probes} & \makecell{0.2713 [0.2175, 0.3275]\\9.0 probes} \\
Random & \makecell{0.1660 [0.1396, 0.1936]\\2.0 probes} & \makecell{0.2214 [0.1896, 0.2548]\\3.0 probes} & \makecell{0.3261 [0.2853, 0.3671]\\6.0 probes} & \makecell{0.4236 [0.3771, 0.4716]\\12.0 probes} \\
\midrule
Global Best (static reference) & \makecell{0.4700 [0.4138, 0.5275]\\0.0 probes} & \makecell{0.4700 [0.4125, 0.5300]\\0.0 probes} & \makecell{0.4700 [0.4113, 0.5275]\\0.0 probes} & \makecell{0.4700 [0.4125, 0.5275]\\0.0 probes} \\
\bottomrule
\end{tabular}
\vspace{2pt}
\begin{minipage}{0.96\textwidth}
\footnotesize
Each cell gives success with a 95\% condition-cluster interval and mean probes.
Semantic Top-1 retrieval uses at most three repeated probes of its retrieved expert; Top-M methods saturate when their fixed shortlist has consumed the three available non-scored trials per expert.
At the $B=12$ cap, RouterVLA uses 12 probes while the semantic shortlist uses nine, so that column is not an exact cost match.
Global Best uses no target-condition probes, so its contrast with routed methods measures conditional deployment against one expert rather than a cost-matched router improvement.
\end{minipage}
\end{table}

\begin{table}[H]
\centering
\caption{Sensitivity of the paired RouterVLA-versus-semantic-shortlist effect to the statistical estimand.}
\label{tab:inference-sensitivity}
\scriptsize
\setlength{\tabcolsep}{2.4pt}
\begin{tabular}{@{}lrrrr@{}}
\toprule
Estimand & $B=2$ & $B=3$ & $B=6$ & $B=12$ \\
\midrule
Condition-weighted, condition cluster & +0.33 [-1.30, +1.98] & -0.56 [-2.64, +1.34] & +1.64 [+0.35, +3.01] & +2.88 [+0.85, +5.14] \\
Condition-weighted, task cluster & +0.33 [-1.26, +1.99] & -0.56 [-3.22, +1.63] & +1.64 [+0.34, +2.95] & +2.88 [+0.96, +4.98] \\
Task-balanced & -0.26 [-2.67, +1.62] & -0.14 [-1.68, +1.31] & +1.11 [+0.16, +2.29] & +1.80 [+0.54, +3.34] \\
Task-balanced hierarchical & -0.26 [-2.71, +1.76] & -0.14 [-1.91, +1.62] & +1.11 [-0.04, +2.47] & +1.80 [+0.30, +3.69] \\
\bottomrule
\end{tabular}
\vspace{2pt}
\begin{minipage}{0.94\textwidth}
\footnotesize
Entries are percentage-point differences with 95\% bootstrap intervals.
The primary condition-weighted analysis matches the benchmark use case; task-balanced variants give equal weight to the 54 task templates.
At $B=12$, RouterVLA uses 12 probes and the semantic shortlist uses nine; this column is a paired nominal-cap contrast, not an exact cost match.
\end{minipage}
\end{table}

\Cref{tab:full-routing-matrix,tab:inference-sensitivity} provide the complete budget sweep and alternate estimands.
The condition-weighted and task-clustered intervals support the exact cost-matched effect at $B=6$ and the direction of the nominal-cap contrast at $B=12$.
Equal weighting across task templates reduces the estimated effect because several templates contribute only one evaluation condition.
The most conservative task-balanced hierarchical interval is directionally positive but includes zero at $B=6$ and excludes zero at $B=12$.
Accordingly, the main paper claims generalization to held-out benchmark conditions and does not claim unseen task-template transfer.

\subsection{Acquisition Controls}

\begin{figure}[H]
  \centering
  \includegraphics[width=0.47\textwidth]{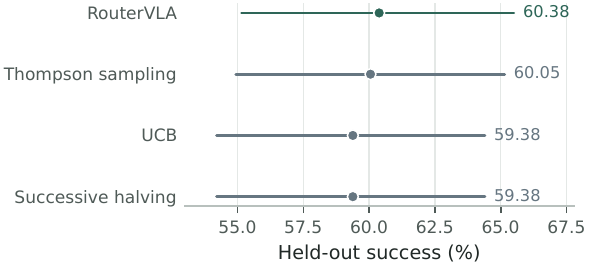}
  \caption{Matched acquisition gauntlet at $B=6$.
  Error bars are 95\% condition-cluster intervals for method success; every rule uses a per-expert probe cap of 2.}
  \label{fig:gauntlet}
\end{figure}

\begin{table}[H]
\centering
\caption{Matched active-acquisition controls under the split-clean prior at budget $B=6$.}
\label{tab:acquisition-gauntlet}
\small
\setlength{\tabcolsep}{6pt}
\begin{tabular}{@{}lcc@{}}
\toprule
Method & SR [95\% CI] & RouterVLA $-$ method (pp) \\
\midrule
\textbf{RouterVLA} & \textbf{0.6038 [0.5515, 0.6550]} & n/a \\
Thompson sampling & 0.6005 [0.5493, 0.6515] & +0.33 [-0.55, +1.18] \\
UCB & 0.5938 [0.5420, 0.6440] & +1.00 [-0.75, +2.55] \\
Successive halving & 0.5938 [0.5420, 0.6440] & +1.00 [-0.75, +2.55] \\
\bottomrule
\end{tabular}
\end{table}

\Cref{fig:gauntlet,tab:acquisition-gauntlet} compare the active allocation rules under matched inputs.
All four active acquisition methods use the same split-clean prior, valid candidate roster, and per-expert probe cap of 2 in this control.
Their paired intervals do not resolve the differences, which supports the paper's emphasis on the prior and trial-disjoint protocol rather than on a universal advantage of one generic bandit heuristic.

\subsection{Summary-Only Confidence-Stopping Diagnostics}

\begin{table}[H]
\centering
\caption{Aggregate confidence-stopping diagnostics from a separate $B=6$ replay.}
\label{tab:stopping-diagnostics}
\small
\setlength{\tabcolsep}{6pt}
\begin{tabular}{@{}lccc@{}}
\toprule
Method & SR [95\% marginal CI] & Mean probes & Artifact scope \\
\midrule
Same-prior stop & \textbf{0.6205 [0.5648, 0.6751]} & 5.685 & Summary only \\
Training-prior stop & 0.5655 [0.5086, 0.6226] & 5.206 & Summary only \\
\bottomrule
\end{tabular}
\end{table}

\Cref{tab:stopping-diagnostics} preserves two aggregate rows from a separate $B=6$ stopping run.
The same-prior rule has the highest point estimate in the package, 62.05\%, with 5.69 mean probes; the training-prior rule reaches 56.55\% with 5.21 probes.
Neither row appears in the primary manifest or the shipped decision-level routing predictions.
Without matched records, we cannot compute the paired contrasts used for the primary claims, so these rows remain descriptive and do not enter the budget sweep.

\subsection{Candidate-Level Onboarding Evidence}

\begin{table}[H]
\centering
\caption{Calibration breadth and one-at-a-time commissioning diagnostics for onboarding candidates.}
\label{tab:onboarding-candidates}
\small
\setlength{\tabcolsep}{5.5pt}
\begin{tabular}{@{}rlrrrc@{}}
\toprule
Floor rank & Candidate & Calibration floor & One-at-a-time gain & Added cells & Useful \\
\midrule
\textbf{1} & $\pi_{0.5}$ & \textbf{32.35\%} & \textbf{+14.83 pp} & \textbf{70} & \textbf{yes} \\
\textbf{2} & \textbf{X-VLA} & \textbf{30.77\%} & \textbf{+13.14 pp} & \textbf{62} & \textbf{yes} \\
\textbf{3} & \textbf{OpenVLA-OFT} & \textbf{16.18\%} & \textbf{+8.90 pp} & \textbf{42} & \textbf{yes} \\
\textbf{4} & \textbf{Xiaomi Robotics-0} & \textbf{5.56\%} & \textbf{+0.85 pp} & \textbf{4} & \textbf{yes} \\
5 & GR00T N1 & 1.39\% & +2.12 pp & 10 & yes \\
6 & CogACT & 0.00\% & +0.00 pp & 0 & no \\
7 & LaST-R1 long & 0.00\% & +1.27 pp & 6 & yes \\
8 & LaST-R1 spatial & 0.00\% & +0.00 pp & 0 & no \\
9 & LingBot-VLA & 0.00\% & +0.00 pp & 0 & no \\
10 & MergeVLA goal & 0.00\% & +2.12 pp & 10 & yes \\
\textbf{11} & \textbf{MergeVLA object} & \textbf{0.00\%} & \textbf{+7.20 pp} & \textbf{34} & \textbf{yes} \\
12 & QwenVLA object & 0.00\% & +0.00 pp & 0 & no \\
13 & RDT & 0.00\% & +0.00 pp & 0 & no \\
14 & StarVLA & 0.00\% & +0.85 pp & 4 & yes \\
15 & VLA-Adapter long & 0.00\% & +0.00 pp & 0 & no \\
16 & VLA-Adapter object & 0.00\% & +0.00 pp & 0 & no \\
\bottomrule
\end{tabular}
\vspace{2pt}
\begin{minipage}{0.94\textwidth}
\footnotesize
The floor is the minimum of the four suite-specific success rates on calibration conditions; zero success on any suite therefore gives a zero floor, and expert ID breaks ties.
Bold rows form the coverage-qualified set: each has positive marginal coverage, and calibration mean ranks the qualified candidates.
One-at-a-time gain holds the 12-expert base fixed, matching the qualification in Equation~\ref{eq:onboarding}.
\end{minipage}
\end{table}

\Cref{tab:onboarding-candidates} separates three quantities: the calibration floor, marginal coverage against the fixed base pool, and membership in the final coverage-qualified set.
Nine of the 16 candidates add positive oracle coverage on the commissioning partition.
The five positive-floor candidates are all useful, and the floor-to-one-at-a-time-coverage rank correlation is $\rho=0.731$.
Four other useful specialists receive a zero floor because each has no recorded calibration success in at least one suite.
The calibration-mean rule selects five candidates, and all five pass positive marginal-coverage qualification.
Qualification therefore leaves the current set unchanged: \pio, X-VLA, OpenVLA-OFT, Xiaomi Robotics-0, and MergeVLA object.
This match is specific to the current roster; marginal coverage still tests whether each candidate solves a base-pool failure.
The minimum-suite floor selects GR00T N1 instead of MergeVLA object because the latter has zero calibration success outside its specialty.
We retain the floor only as a breadth diagnostic; it neither determines the main pool nor supplies an accuracy claim.
The marginal gains in the table hold the base pool fixed, matching the qualification rule in \cref{eq:onboarding}.

\begin{table}[H]
\centering
\caption{Twenty random five-expert expansions evaluated with the same routed policy.}
\label{tab:random-pools}
\scriptsize
\setlength{\tabcolsep}{3.5pt}
\begin{tabularx}{0.96\textwidth}{@{}rXcc@{}}
\toprule
Draw & Added experts & SR & Probes \\
\midrule
0 & groot, mvobject, oft, \pio, vla\_adapter\_object & 0.5540 & 6.000 \\
1 & lastr1\_spatial, mvgoal, mvobject, starvla, xvla & 0.5170 & 5.865 \\
2 & cogact, lastr1\_long, lastr1\_spatial, qwenvla\_object, starvla & 0.3455 & 6.000 \\
3 & groot, lingbot, mvgoal, qwenvla\_object, xvla & 0.4645 & 6.000 \\
4 & lastr1\_long, mvobject, \pio, rdt, starvla & 0.5012 & 6.000 \\
5 & groot, qwenvla\_object, rdt, xiaomi, xvla & 0.4788 & 6.000 \\
6 & cogact, groot, lastr1\_long, lingbot, xvla & 0.4597 & 6.000 \\
7 & lastr1\_long, lingbot, mvgoal, mvobject, vla\_adapter\_object & 0.4183 & 5.850 \\
8 & lastr1\_long, mvobject, oft, \pio, starvla & 0.5517 & 6.000 \\
9 & groot, lastr1\_spatial, qwenvla\_object, xiaomi, xvla & 0.4830 & 6.000 \\
10 & cogact, mvobject, rdt, xiaomi, xvla & 0.5242 & 5.865 \\
11 & lingbot, rdt, starvla, xiaomi, xvla & 0.4753 & 5.865 \\
12 & cogact, groot, lastr1\_spatial, qwenvla\_object, xiaomi & 0.3750 & 6.000 \\
13 & lingbot, mvobject, starvla, vla\_adapter\_object, xvla & 0.5030 & 6.000 \\
14 & lingbot, mvobject, oft, qwenvla\_object, xiaomi & 0.4780 & 6.000 \\
15 & lingbot, mvobject, \pio, rdt, vla\_adapter\_long & 0.5025 & 5.865 \\
16 & cogact, lingbot, \pio, vla\_adapter\_object, xvla & 0.5547 & 6.000 \\
17 & groot, mvgoal, mvobject, \pio, xvla & 0.5715 & 6.000 \\
18 & lastr1\_long, lingbot, mvgoal, oft, rdt & 0.4542 & 6.000 \\
19 & groot, \pio, vla\_adapter\_long, xiaomi, xvla & 0.5477 & 6.000 \\
\midrule
Selected & \pio, X-VLA, OpenVLA-OFT, Xiaomi Robotics-0, MergeVLA object & \textbf{0.6096} & 6.000 \\
\bottomrule
\end{tabularx}
\end{table}

\Cref{tab:random-pools} lists all sampled expansions.
The 20 random expansions range from 34.55\% to 57.15\% held-out routed success.
The selected pool reaches 60.96\%, above every sampled random expansion.
Its $12.16\pp$ advantage over the random mean is descriptive across these 20 draws; the between-draw standard deviation is $5.94\pp$.
The selected pool's 60.96\% point estimate is $0.01\pp$ above the full roster's 60.95\% point estimate.
These existing replays do not form a paired equivalence test.

\subsection{Full Expert Roster}

\begin{table}[H]
\centering
\caption{Complete 28-expert roster and suite-level success in the trial-disjoint LIBERO-Plus ledger.}
\label{tab:expert-pool}
\scriptsize
\setlength{\tabcolsep}{3.2pt}
\renewcommand{\arraystretch}{1.04}
\begin{tabularx}{\textwidth}{@{}l>{\raggedright\arraybackslash}Xlccccc@{}}
\toprule
Expert & Source & Role & Long & Goal & Object & Spatial & Overall \\
\midrule
$\pi_{0.5}$ & \textbf{\citep{black2024pi0}} & \textbf{Selected} & \textbf{0.384} & \textbf{0.241} & \textbf{0.644} & \textbf{0.759} & \textbf{0.507} \\
\textbf{X-VLA} & \textbf{\citep{zheng2025xvla}} & \textbf{Selected} & \textbf{0.359} & \textbf{0.206} & \textbf{0.397} & \textbf{0.688} & \textbf{0.412} \\
\textbf{OpenVLA-OFT} & \textbf{\citep{kim2025openvlaoft}} & \textbf{Selected} & \textbf{0.188} & \textbf{0.237} & \textbf{0.422} & \textbf{0.444} & \textbf{0.323} \\
DB-CogACT & \citep{li2024cogact} & Base & 0.197 & 0.116 & 0.431 & 0.497 & 0.310 \\
\textbf{Xiaomi Robotics-0} & \textbf{\citep{cai2026xiaomi}} & \textbf{Selected} & \textbf{0.156} & \textbf{0.066} & \textbf{0.331} & \textbf{0.469} & \textbf{0.255} \\
\textbf{MergeVLA object} & \textbf{\citep{fu2025mergevla}} & \textbf{Selected} & \textbf{0.000} & \textbf{0.000} & \textbf{0.613} & \textbf{0.000} & \textbf{0.153} \\
StarVLA & \citep{zhou2026starvla} & Candidate & 0.022 & 0.047 & 0.122 & 0.384 & 0.144 \\
MergeVLA spatial & \citep{fu2025mergevla} & Base & 0.000 & 0.000 & 0.000 & 0.569 & 0.142 \\
GR00T N1 & \citep{bjorck2025groot} & Candidate & 0.009 & 0.069 & 0.256 & 0.228 & 0.141 \\
VLA-Adapter spatial & \citep{wang2025vlaadapter} & Base & 0.000 & 0.000 & 0.000 & 0.559 & 0.140 \\
VLANeXt & \citep{wu2026vlanext} & Base & 0.066 & 0.113 & 0.031 & 0.206 & 0.104 \\
OpenVLA-LoRA & \citep{kim2024openvla} & Base & 0.003 & 0.007 & 0.100 & 0.194 & 0.078 \\
OpenVLA & \citep{kim2024openvla} & Base & 0.013 & 0.003 & 0.050 & 0.184 & 0.062 \\
VLA-Adapter object & \citep{wang2025vlaadapter} & Candidate & n/a & 0.000 & 0.184 & 0.000 & 0.061 \\
LaST-R1 object & \citep{chen2026lastr1} & Base & 0.000 & 0.000 & 0.188 & 0.000 & 0.047 \\
MergeVLA long & \citep{fu2025mergevla} & Base & 0.156 & 0.000 & 0.000 & 0.000 & 0.039 \\
MergeVLA goal & \citep{fu2025mergevla} & Candidate & 0.000 & 0.141 & 0.000 & 0.000 & 0.035 \\
LaST-R1 long & \citep{chen2026lastr1} & Candidate & 0.078 & 0.003 & 0.000 & 0.009 & 0.023 \\
VLA-Adapter goal & \citep{wang2025vlaadapter} & Base & 0.000 & 0.075 & 0.000 & 0.000 & 0.019 \\
VLA-Adapter long & \citep{wang2025vlaadapter} & Candidate & 0.041 & 0.000 & 0.000 & 0.000 & 0.010 \\
LaST-R1 goal & \citep{chen2026lastr1} & Base & 0.000 & 0.034 & 0.000 & 0.000 & 0.009 \\
QwenVLA object & source ledger & Candidate & 0.000 & 0.000 & 0.000 & 0.003 & 0.001 \\
LaST-R1 spatial & \citep{chen2026lastr1} & Candidate & 0.000 & 0.003 & 0.000 & 0.000 & 0.001 \\
CogACT & \citep{li2024cogact} & Candidate & 0.000 & 0.000 & 0.000 & 0.000 & 0.000 \\
LingBot-VLA & \citep{wu2026pragmatic} & Candidate & 0.000 & 0.000 & n/a & n/a & 0.000 \\
SmolVLA & \citep{shukor2025smolvla} & Base & 0.000 & 0.000 & 0.000 & 0.000 & 0.000 \\
QwenVLA spatial & source ledger & Base & 0.000 & 0.000 & 0.000 & 0.000 & 0.000 \\
RDT & \citep{liu2024rdt} & Candidate & 0.000 & 0.000 & 0.000 & 0.000 & 0.000 \\
\bottomrule
\end{tabularx}
\vspace{2pt}
\begin{minipage}{0.95\textwidth}
\footnotesize
n/a denotes an unevaluated suite, whereas 0.000 denotes evaluated zero success.
``Base'' marks the 12 inherited operational experts; bold rows form the coverage-qualified calibration-mean top five.
\end{minipage}
\end{table}

\Cref{tab:expert-pool} identifies the inherited base, selected additions, and remaining candidates.
The roster combines broad models with suite-specialized adapters and modular heads.
The top overall expert is not uniformly best across suites, and several lower-average specialists solve cells that broader models miss.
Onboarding looks for those specialists, and routing assigns them where they help.

\subsection{Three Cobblers Beat a Mastermind}
\label{sec:legacy-weak-router}

\paragraph{A separate restricted-pool experiment.}
We also tested a deliberately different question from the main study: can condition-level routing combine individually weaker policies into a system that outperforms the strongest fixed open-source generalist in the evaluated snapshot?
This experiment is self-contained in the supplement.
Its records, splits, features, and estimates do not enter the main routing, onboarding, or pool-expansion results.

\paragraph{Experimental setup.}
The experiment used a 320-variant LIBERO-Plus snapshot spanning four suites and 14 specialist policies from seven architecture families.
The candidate set explicitly excluded $\pi_{0.5}$~\citep{physicalintelligence2025pi05}; depending on the held-out suite, the router selected among 5--14 policies that were individually weaker than that fixed reference.
Each candidate was represented by a 12-dimensional cross-specialist-relative (CSR) profile aggregated from four rollouts: step-count distribution summaries, duration statistics, and termination behavior, z-scored across the available pool for the same variant.
A $12\!\rightarrow\!64\!\rightarrow\!32\!\rightarrow\!|\E|$ MLP was trained with the reinforce policy-gradient estimator for 1{,}500 epochs under leave-one-suite-out evaluation.
We used five seeds and 10{,}000 percentile bootstrap resamples.
This setup is intentionally distinct from the main paper's development/commissioning/evaluation partitions, total-budget acquisition rule, and marginal-coverage onboarding protocol.

\begin{table}[H]
  \centering
  \caption{Restricted-pool routing with $\pi_{0.5}$ excluded. Gains in this table are paired against the best fixed expert inside the restricted weak pool; the direct comparison with the separately reported fixed $\pi_{0.5}$ reference is stated in the text.}
  \label{tab:restricted-pool}
  \begin{tabular}{lcc}
    \toprule
    Held-out split & Gain (pp) & 95\% CI \\
    \midrule
    LIBERO Object  & $+13.11$ & $[+8.18,+18.11]$ \\
    LIBERO Spatial & $+13.75$ & $[+10.94,+16.50]$ \\
    LIBERO Long    & $+13.88$ & $[+9.62,+18.12]$ \\
    LIBERO Goal    & $+23.32$ & $[+19.88,+26.66]$ \\
    \midrule
    Joint          & $\mathbf{+15.91}$ & $\mathbf{[+11.47,+20.41]}$ \\
    \bottomrule
  \end{tabular}
\end{table}

\paragraph{Weaker models, routed together, surpassed $\pi_{0.5}$.}
In this benchmark snapshot, $\pi_{0.5}$ was the strongest fixed open-source reference at 50.7\% joint success.
The router never had access to $\pi_{0.5}$: it selected only among the weaker candidate policies, yet reached 69.1\% joint success, a direct improvement of 18.4 percentage points over the fixed reference.
The restricted-pool upper reference was 69.4\%, placing the learned router only 0.3 points below the best success available from that weak pool.
Within this experimental setup, the routed weak pool therefore produced the strongest result among the evaluated open models.

\paragraph{Baseline accounting.}
The $+15.91\pp$ joint gain in \cref{tab:restricted-pool} and the $+18.4\pp$ direct comparison use different fixed references.
The table reports paired gains over the best fixed expert \emph{inside the restricted pool}; the direct comparison contrasts the routed 69.1\% result with the separately reported 50.7\% $\pi_{0.5}$ result.
The first quantifies conditional value within the weak pool, while the second states the ``three cobblers beat a mastermind'' result directly.

\paragraph{Relation to the main study.}
This auxiliary experiment uses a different data snapshot, exhaustive four-rollout CSR profiles, cross-specialist normalization, reinforce training, and a different candidate definition.
It answers a complementary question rather than validating the present method: the main paper studies split-clean, total-budget commissioning and expert onboarding, whereas this experiment isolates whether routing can compose weaker models into a system stronger than the strongest fixed open-source reference.

\subsection{Representative Simulator Frames}

\begin{figure}[H]
  \centering
  \begin{minipage}[t]{0.235\textwidth}
    \centering
    \includegraphics[width=\linewidth]{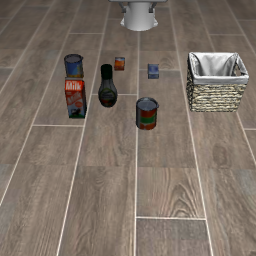}
    \small Object suite
  \end{minipage}
  \hfill
  \begin{minipage}[t]{0.235\textwidth}
    \centering
    \includegraphics[width=\linewidth]{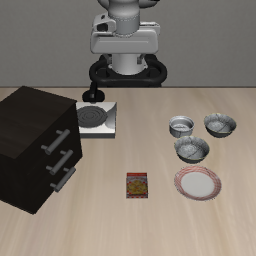}
    \small Spatial suite
  \end{minipage}
  \hfill
  \begin{minipage}[t]{0.235\textwidth}
    \centering
    \includegraphics[width=\linewidth]{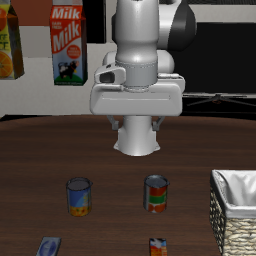}
    \small Long-horizon suite
  \end{minipage}
  \hfill
  \begin{minipage}[t]{0.235\textwidth}
    \centering
    \includegraphics[width=\linewidth]{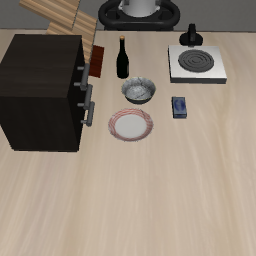}
    \small Goal-conditioned suite
  \end{minipage}
  \caption{Representative raw LIBERO-Plus simulator frames.
  The images provide visual context and are not inputs to the present text-and-rollout router.}
  \label{fig:raw-frames}
\end{figure}
\Cref{fig:raw-frames} provides visual context for the task suites; the router does not consume these images.

\end{document}